\documentclass[runningheads]{llncs}

\usepackage{eccv}

\usepackage{eccvabbrv}

\usepackage{graphicx}
\usepackage{booktabs}

\usepackage[accsupp]{axessibility}  

\usepackage[pagebackref,breaklinks,colorlinks,citecolor=eccvblue]{hyperref}

\usepackage{orcidlink}

\usepackage{multirow}

\newcommand{\RVT}{RVT-2}

\begin{document}

\title{Energy-Regularized Imitation Learning for Force- and Work-Aware Robotic Manipulation}
\titlerunning{Energy-Regularized Imitation Learning}

\author{Toshiki Otani, Hiromu Taketsugu, Norimichi Ukita}
\authorrunning{T. Otani, et al.}
\institute{Toyota Technological Institute, Japan}

\maketitle

\begin{abstract}
This paper studies energy-aware manipulation as a physically grounded learning problem. We define a joint-space mechanical-work proxy from joint torque and angular displacement, and train a differentiable energy predictor that estimates this work from robot states and actions. The predictor converts a non-differentiable simulator-side physical quantity into a differentiable regularizer for fine-tuning a pretrained manipulation policy. We instantiate the framework with RVT-2 on RLBench and evaluate 12 manipulation tasks involving object contact, articulated motion, placement, pushing, and sweeping. The proposed fine-tuning reduces the average mechanical work from 208.8~J to 204.4~J (i.e., 2.1\% reduction), while the mean task success rate also increases slightly from 86.2\% to 86.9\%. These results show that work-aware policy optimization can suppress physically inefficient motion without requiring an explicit differentiable dynamics model.
\end{abstract}

\begin{figure}[t]
    \begin{center}
        \includegraphics[width=0.95\linewidth]{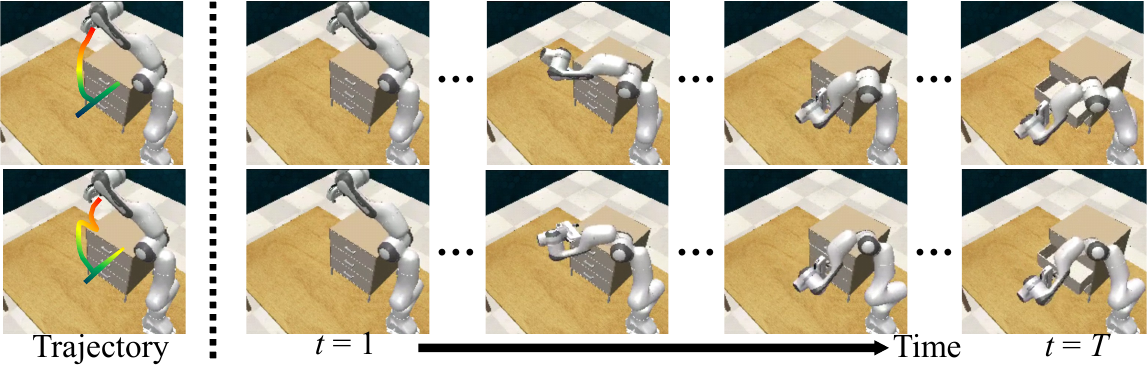}
        \vspace*{-2mm}
        \caption{A manipulation task can be completed with either (Upper) a short, physically efficient trajectory or (Lower) a redundant trajectory involving unnecessary contact.}
        \label{fig:teaser}
    \end{center}
    \vspace*{-4mm}
\end{figure}

\section{Introduction}
\label{sec:intro}

Imitation learning, including PerAct~\cite{peract}, Q-Transformer~\cite{qtransformer}, and \RVT{}~\cite{rvt2}, can learn object manipulation from demonstrations.
Their training objectives are usually defined around imitation accuracy
without explicitly evaluating the physical effort required by the executed motion.
A policy may therefore complete a task through redundant accelerations, unnecessarily large joint motion, or avoidable contact with the environment.
Figure~\ref{fig:teaser} shows this contrast.

This issue is relevant to force-grounded articulated manipulation, where task completion alone is insufficient to characterize the physical quality of a robot motion. In contact-rich settings, force, torque, and mechanical work are central quantities because they determine how much physical effort is exerted on the robot, the object, and the surrounding environment. In this paper, we address this problem from the policy-learning side by using mechanical work as an additional training signal for a manipulation policy.

Directly using mechanical work as a training loss is difficult because torque and contact computations are produced inside physics engines~\cite{coppeliaSim,mujoco}. These computations are not differentiable functions of a high-level neural policy. In real robots, dynamics parameters and contact states are also difficult to identify. We therefore introduce a learned surrogate predictor that converts simulator-side work measurements into a differentiable objective.

We propose energy-regularized imitation learning. 
First, we compute action-level work by accumulating joint-space work over the simulator steps required to execute each key action. Second, we train an energy predictor that approximates this quantity from robot states and actions. Third, we freeze the predictor and fine-tune a pretrained policy network with an additional energy loss.

Our contributions are threefold. (1) We formulate manipulation energy consumption as mechanical work computed from joint torque and angular displacement, aligning policy learning with force-, torque-, and work-aware manipulation. (2) We introduce a differentiable energy predictor and integrate it into a pretrained imitation-learning policy as a regularizer. (3) We evaluate the method on 12 RLBench tasks and show an average 2.1\% reduction in mechanical work with no degradation in average success rate.


\section{Related Work}
\label{sec:related}

\paragraph{Imitation learning for visual manipulation.}
Behavioral cloning learns a policy from expert demonstrations~\cite{bc}. It is simple and data efficient, yet suffers from distribution shift when the policy visits states outside that distribution~\cite{dagger}. Recent methods~\cite{peract,read,rvt2,DBLP:journals/ijon/ObaU25,DBLP:journals/corr/abs-2504-19652,DBLP:conf/icra/ObaU23} improve policy expressivity through diffusion models, transformers, 3D scene representations, and multi-view observations. These methods mainly optimize action prediction and task success. Recent reinforcement-learning methods incorporate torque- and mechanical-energy penalties into policy optimization for reaching, box pushing, mobile manipulation, and articulated-object manipulation~\cite{otto2023deep,fu2023deep,tao2026energy,deniz2026energy}. Unlike these RL methods, our method learns a differentiable surrogate of simulator-computed joint-space work and uses it to fine-tune a pretrained visual imitation-learning policy.


\paragraph{Force-, torque-, and work-aware manipulation.}
Classical robot control uses dynamics models, torque limits, and trajectory optimization to reduce control cost or mechanical effort~\cite{rrtconnect}. In learned manipulation, however, exact dynamics are difficult to obtain for contact-rich tasks. Existing benchmarks such as RLBench~\cite{rlbench} and Meta-World~\cite{metaworld} mainly evaluate task success. We instead use mechanical work as an additional optimization target.


\paragraph{Surrogate physical models.} When a physical quantity is computed inside a simulator, it is often unavailable as a differentiable function of a neural policy. Neural surrogate models can approximate such quantities and provide gradients for policy optimization~\cite{surrogate_complex_physics,surrogate_neural_architecture,difftaichi}. In our work, the surrogate predicts manipulator work from robot states and actions, without simulating full contact dynamics.


\section{Method}
\label{sec:method}

\subsection{Mechanical Work as the Energy Target}

We define the energy target as a joint-space mechanical-work proxy computed from joint torque and joint displacement. Let $J$ be the number of robot joints and $T$ be the number of control steps. At step $t$, the simulator provides the joint torque vector $\boldsymbol{\tau}_t \in \mathbb{R}^{J}$ and joint angle vector $\boldsymbol{q}_t \in \mathbb{R}^{J}$. The per-step mechanical work $e_t$ and the trajectory-level work $E(\boldsymbol{\tau}_{1:T},\boldsymbol{q}_{0:T})$ are defined as follows:
\begin{eqnarray}
    e_t &=& \textstyle\sum_{j=1}^{J} |\tau_{t,j}|\, |q_{t,j}-q_{t-1,j}|,
    \label{eq:stepwork}\\
    E(\boldsymbol{\tau}_{1:T},\boldsymbol{q}_{0:T}) &=& \textstyle\sum_{t=1}^{T} e_t.
    \label{eq:trajwork}
\end{eqnarray}
The policy network predicts key actions at a higher level than the simulator control step.
Let $\mathcal{T}_k$ be the set of simulator steps required to execute key action $k$.
The action-level work is defined as follows:
\begin{equation}
    E_{\mathrm{act}}(k) = \textstyle\sum_{t\in\mathcal{T}_k} e_t.
    \label{eq:actionwork}
\end{equation}

These definitions follow the physical form of work, torque multiplied by angular displacement. 
The proposed quantity is an absolute joint-space work proxy for manipulator effort, rather than the net mechanical work transferred to the object or the electrical energy consumed by the actuators. It is complementary to contact-gated work metrics, which integrate force along distance during contact. We use this joint-space formulation because RLBench provides robot joint states and simulator-side torques for all tasks.


Although this work measure is physically meaningful, it is not directly usable as a gradient-based loss for a high-level policy. The torque values are computed inside the simulator after executing actions. Contacts, collision handling, and low-level control introduce discontinuities, and the simulator is not part of the neural computation graph. We therefore learn a differentiable approximation of the action-level work in Eq.~\eqref{eq:actionwork}, as described in Sec.~\ref{subsec:energy_predictor}.

\begin{figure}[t]
    \begin{center}
        \includegraphics[width=\linewidth]{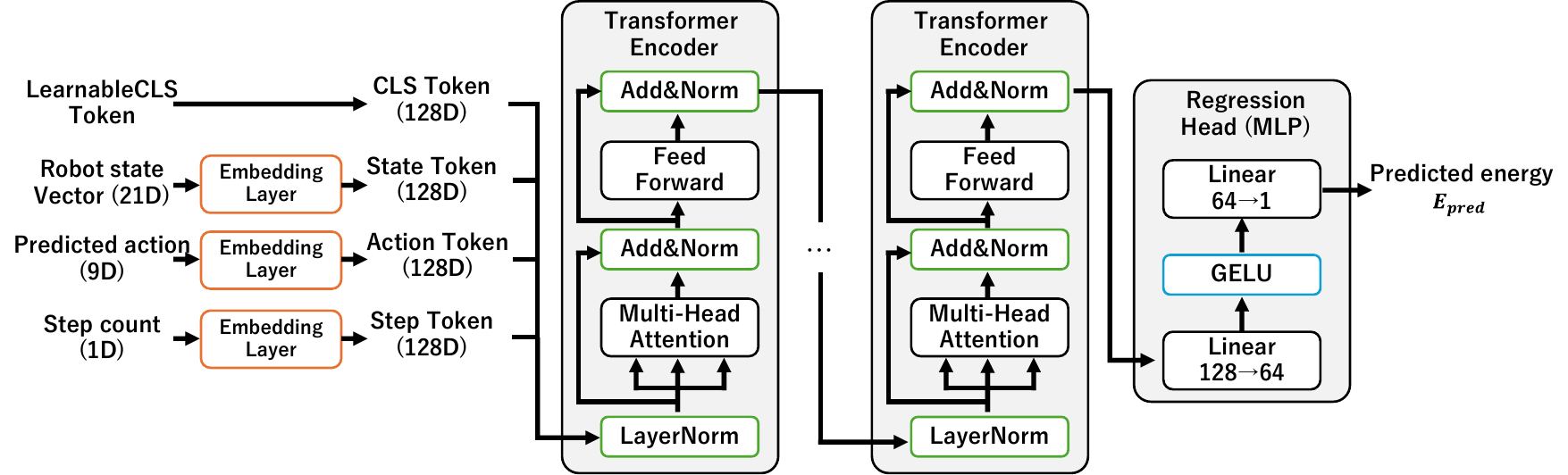}
        \caption{Network architecture of our energy prediction network.}
        \label{fig:energy_predictor}
    \end{center}
\end{figure}

\subsection{Differentiable Energy Predictor}
\label{subsec:energy_predictor}

Let $\mathbf{s}_k$ denote the robot state at key action $k$, and let
$\mathbf{a}_k$ denote an input key action. We train an energy prediction
network parameterized by $\phi$ (denoted by $g_{\phi}$):
\begin{equation}
    \hat{E}_{\phi}=g_{\phi}(\mathbf{s}_k,\mathbf{a}_k)
    \label{eq:predictor}
\end{equation}
$g_{\phi}$ estimates the action-level work in Eq.~\eqref{eq:actionwork}.
The predictor is trained on demonstration key actions, using simulator-computed
work labels obtained during action execution. It does not take simulator torque
as input; torque and joint displacement are used only to compute the supervised
work label. Figure~\ref{fig:energy_predictor} shows the predictor architecture.
The predictor input consists of a 21-D robot-state vector, a 9-D action, and a 1-D step-count feature. Specifically, the 21-D robot-state vector consists of the angles and angular velocities of seven joints (14-D), the end-effector xyz position (3-D), and the end-effector quaternion (4-D). The 9-D action consists of the target xyz position (3-D), target quaternion (4-D), gripper open/close command (1-D), and collision flag (1-D). 
The 1-D step-count feature denotes the current task-step index and provides
task-progress information to the non-recurrent predictor.
This predictor is trained with RAdamScheduleFree~\cite{radamschedulefree}, a learning rate of $10^{-3}$, weight decay $10^{-4}$, batch size 1024, and early stopping.

We use a Huber loss to make $\hat{E}_{\phi}$ robust to rare high-energy episodes:
\begin{equation}
\mathcal{L}_{\mathrm{pred}}(\phi)
=
\mathrm{Huber}
\left(
g_{\phi}(\mathbf{s}_k,\mathbf{a}_k),
E_{\mathrm{act}}(k)
\right).
\label{eq:pred_loss}
\end{equation}

Before training the predictor, we filter atypical successful steps using
HDBSCAN~\cite{hdbscan}. For each task, clustering is performed in a
two-dimensional space of arm-motion magnitude and measured action-level work.
Atypical successful behaviors, such as re-grasping, repeated approaches,
accidental contact, or temporary stuck states, often deviate from the normal
motion--work correlation by producing high torque with small motion. Samples in
low-density outlier regions are removed before predictor training, keeping the
predictor focused on representative successful motions.

\subsection{Energy-Regularized Fine-Tuning}

\begin{figure}[t]
    \begin{center}
        \includegraphics[width=\linewidth]{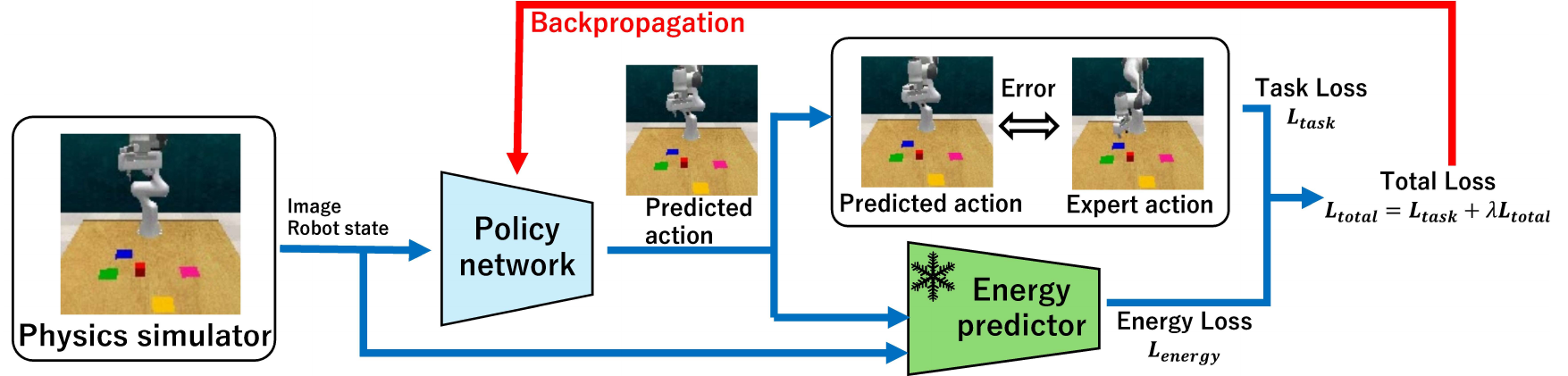}
        \caption{Overview of energy-regularized fine-tuning.}
        \label{fig:overview}
    \end{center}
\end{figure}

We integrate $g_{\phi}$ into policy-network fine-tuning. 
Figure~\ref{fig:overview} shows the overall fine-tuning framework.
Let $\mathcal{L}_{\mathrm{task}}$ be the original imitation-learning loss that supervises the predicted action $\hat{\mathbf{a}}_k$ with the expert action $\mathbf{a}^{*}_k$ from the demonstration.
During fine-tuning, $g_{\phi}$ is frozen and only the manipulation policy parameters $\theta$ are updated. At each training step, the policy network predicts an action $\hat{\mathbf{a}}_k=\pi_\theta(\mathbf{o}_k)$, which is supervised by the corresponding expert action $\mathbf{a}^{*}_k$ from the demonstration.
The same predicted action, together with the robot state, is fed to the frozen $g_{\phi}$ to compute the work regularization term.

The energy loss is the min-max normalized predicted work using Eq.~(\ref{eq:predictor}):
\begin{equation}
    \mathcal{L}_{\mathrm{energy}}
    =
    \frac{
    \hat{E}_{\phi}-E_{\min}
    }{
    E_{\max}-E_{\min}
    }.
    \label{eq:energy_loss}
\end{equation}
where $E_{\min}$ and $E_{\max}$ denote fixed constants computed once as the minimum and maximum of $E_{\mathrm{act}}(k)$ over the action-level work labels in the energy-predictor training set; they are kept fixed during policy fine-tuning.
The total fine-tuning loss is defined with Eq.~(\ref{eq:energy_loss}):
\begin{equation}
    \mathcal{L}_{\mathrm{total}}(\theta)
    =
    \mathcal{L}_{\mathrm{task}}(\hat{\mathbf{a}}_k,\mathbf{a}^{*}_k)
    +
    \lambda
    \mathcal{L}_{\mathrm{energy}},
    \label{eq:finetune}
\end{equation}
where $\lambda$ denotes a weight that controls the tradeoff between action imitation and work reduction. We use $\lambda=10$ as the main setting. Fine-tuning uses LAMB~\cite{lamb}, learning rate $10^{-4}$, a batch size of 2 per GPU, cosine annealing with warmup~\cite{cosine}, and early stopping. Since the predictor is frozen, the policy cannot reduce the loss by changing the physical model. It must instead produce actions that the predictor estimates as lower work.


\section{Experiments}
\label{sec:experiments}

\subsection{Experimental Setup}
We evaluate on 12 RLBench tasks~\cite{rlbench}: Close Jar, Light Bulb In, Open Drawer, Place Shape in Shape Sorter, Place Wine at Rack Location, Push Buttons, Put Item in Drawer, Reach and Drag, Stack Blocks, Stack Cups, Sweep to Dustpan of Size, and Turn Tap. These tasks cover articulated objects, placing, button interaction, reaching, stacking, sweeping, and object rearrangement.
We use \RVT{} as the baseline policy and report task success rate and mechanical work in Joules. For evaluation, Work[J] denotes the trajectory-level work in Eq.~\eqref{eq:trajwork}, computed after each policy rollout and averaged over successful episodes only.

\subsection{Energy Predictor Accuracy}
Table~\ref{tab:predictor} reports the prediction accuracy of the energy predictor. The predictor achieves an average mean absolute error (MAE) of 2.40 J and an average relative error of 7.02\% across 12 tasks. Errors are smallest for button interaction, sweeping, and simple reaching tasks, and larger for articulated-object manipulation and long-horizon placement tasks.
This accuracy is sufficient for regularization because the fine-tuning objective mainly requires a stable low-versus-high work signal rather than exact physical reconstruction at every step.

\begin{table}[t]
\centering
\caption{Energy-predictor accuracy on 12 RLBench tasks.
Rel. denotes MAE divided by the mean ground-truth action-level work.}
\label{tab:predictor}
\scriptsize
\setlength{\tabcolsep}{3pt}
\begin{tabular}{lcc|lcc}
\toprule
Task & MAE[J] & Rel.[\%] & Task & MAE[J] & Rel.[\%] \\
\midrule
Close Jar & 1.221 & 4.05 & Put Item Drawer & 0.862 & 3.16 \\
Light Bulb In & 4.775 & 13.81 & Reach and Drag & 1.178 & 5.72 \\
Open Drawer & 4.136 & 11.71 & Stack Blocks & 1.198 & 5.15 \\
Shape Sorter & 4.787 & 14.59 & Stack Cups & 0.878 & 4.33 \\
Place Wine & 2.864 & 7.47 & Sweep Dustpan & 0.565 & 2.13 \\
Push Buttons & 0.211 & 1.86 & Turn Tap & 6.184 & 10.20 \\
\midrule
\multicolumn{4}{r}{Average} & \textbf{2.40} & \textbf{7.02} \\
\bottomrule
\end{tabular}
\end{table}
\begin{table}[t]
\centering
\caption{Comparison between \RVT{} and the proposed energy-regularized policy ($\lambda=10$). Work is averaged over successful episodes only.}
\label{tab:main}
\scriptsize
\setlength{\tabcolsep}{3pt}
\begin{tabular}{lccccc}
\toprule
\multirow{2}{*}{Task} & \multicolumn{2}{c}{Success[\%]} & \multicolumn{2}{c}{Work[J]} & \multirow{2}{*}{Red.[\%]} \\
\cmidrule(lr){2-3}\cmidrule(lr){4-5}
 & \RVT{} & Ours & \RVT{} & Ours & \\
\midrule
Close Jar & 100 & 100 & 171.11 & 170.71 & 0.2 \\
Light Bulb In & 89 & 88 & 223.02 & 222.22 & 0.4 \\
Open Drawer & 78 & 82 & 131.23 & 131.44 & -0.2 \\
Shape Sorter & 41 & 47 & 260.69 & 219.48 & 15.8 \\
Place Wine & 89 & 91 & 207.00 & 205.84 & 0.6 \\
Push Buttons & 100 & 100 & 115.40 & 115.45 & 0.0 \\
Put Item Drawer & 96 & 93 & 357.04 & 348.78 & 2.3 \\
Reach and Drag & 100 & 100 & 145.12 & 145.02 & 0.1 \\
Stack Blocks & 70 & 74 & 403.39 & 408.15 & -1.2 \\
Stack Cups & 74 & 72 & 238.90 & 240.31 & -0.6 \\
Sweep Dustpan & 100 & 100 & 125.60 & 125.51 & 0.1 \\
Turn Tap & 97 & 96 & 126.50 & 119.92 & 5.2 \\
\midrule
Average & 86.2 & \textbf{86.9} & 208.8 & \textbf{204.4} & \textbf{2.1} \\
\bottomrule
\end{tabular}
\end{table}

\subsection{Energy-Regularized Fine-Tuning}
Table~\ref{tab:main} compares the baseline \RVT{} policy and our energy-regularized fine-tuned policy with $\lambda=10$. The proposed method reduces the average mechanical work from 208.8 J to 204.4 J, corresponding to a 2.1\% reduction, while the average success rate also increases from 86.2\% to 86.9\%. 
This suggests that the energy term acts as a mild efficiency prior without harming task completion.
The reduction is conservative because the policy is already pretrained for high success and the regularization weight is kept moderate.

The energy reduction is large for tasks where the baseline tends to produce redundant motion or unnecessary contact. Shape Sorter reduces work by 15.8\%, Turn Tap by 5.2\%, and Put Item Drawer by 2.3\%. Several tasks preserve success completely while slightly reducing work, including Close Jar, Reach and Drag, and Sweep Dustpan. Some tasks show a negative energy reduction, such as Open Drawer, Stack Blocks, and Stack Cups, indicating that the regularizer does not uniformly improve all manipulation modes. These failures suggest that task-dependent contact strategies and success constraints should be incorporated into future energy-aware losses.

\subsection{Ablation Studies}
\paragraph{Effect of atypical-data filtering.}
Table~\ref{tab:hdbscan} shows the effect of filtering atypical successful trajectories before predictor training. HDBSCAN filtering improves the average predictor MAE from 2.80 J to 2.40 J over the 12 tasks. We further report group averages for representative difficult and simple tasks to examine whether this effect holds across different prediction-error regimes. The difficult group contains the four tasks with the largest filtered predictor errors: Turn Tap, Place Shape in Shape Sorter, Light Bulb In, and Open Drawer. The simple group contains the four tasks with the smallest filtered predictor errors: Push Buttons, Sweep to Dustpan of Size, Put Item in Drawer, and Stack Cups. The baseline policy (i.e., \RVT{}) has a mean work of 208.8 J, as shown in Table~\ref{tab:main}. Fine-tuning with the unfiltered predictor increases it to 214.8 J, whereas fine-tuning with the filtered predictor reduces it to 204.4 J, as shown in Table~\ref{tab:hdbscan}. This supports the importance of estimating energy from representative successful behavior rather than from all successful rollouts indiscriminately.

\begin{table}[t]
\centering
\caption{Ablation of atypical-data filtering over the 12 RLBench tasks. Representative difficult and simple tasks are selected by the largest and smallest filtered predictor errors, respectively.}
\label{tab:hdbscan}
\scriptsize
\setlength{\tabcolsep}{4pt}
\begin{tabular}{lcc|lcc}
\toprule
\multirow{2}{*}{Task group} & \multicolumn{2}{c|}{Predictor MAE[J]} & \multicolumn{2}{c}{Fine-tuned work[J]} \\
\cmidrule(lr){2-3}\cmidrule(lr){4-5}
& Filtered & Unfiltered & Filtered & Unfiltered \\
\midrule
Average over 12 tasks & \textbf{2.40} & 2.80 & \textbf{204.4} & 214.8 \\
Four difficult tasks & 4.97 & 6.04 & 173.3 & 202.7 \\
Four simple tasks & 0.63 & 0.74 & 207.5 & 207.7 \\
\bottomrule
\end{tabular}
\end{table}

\paragraph{Using internal \RVT{} features.}
We also tested an energy predictor that uses internal \RVT{} features in addition to robot state and action features. This improves the average predictor MAE from 2.40 J to 2.30 J. However, the downstream fine-tuning does not improve average work: the mean work becomes 207.8 J with an average success rate of 87.3\%. This contrast indicates that predictor MAE alone is not the only criterion for an effective regularizer. The predictor must also provide gradients that are well aligned with controllable action changes.

\paragraph{Sensitivity to the energy weight.}
The coefficient $\lambda$ controls the tradeoff in Eq.~\eqref{eq:finetune}. Moderate weights can reduce work while preserving task completion for several tasks. Excessively large weights harm success, especially in contact-rich insertion and articulated-object tasks, because the policy can minimize predicted work by avoiding decisive contact or by producing overly conservative actions. In our experiments, $\lambda=10$ offers the clearest average work reduction, while $\lambda=5$ and $\lambda=7$ are often safer for success-critical tasks.

\section{Discussion}
\label{sec:discussion}
Our method does not modify the visual encoder of the base policy. Instead, it adds a physically grounded supervision signal to the action prediction head.
The approach is complementary to visual representation learning and can be applied to visual manipulation policies without changing their perception backbone.

The proposed method connects imitation learning with physically grounded manipulation in two ways. First, it uses a work quantity that depends on joint torque and motion, rather than a purely kinematic smoothness penalty. Smooth motion can still be inefficient when the robot pushes against an object or applies unnecessary torque. Second, our method directly targets policy learning. The predictor is trained from physics simulation and then used as a differentiable energy term, avoiding the need to backpropagate through contact simulation.

The current study has limitations. The work definition is joint-space work and does not explicitly gate object contact. Thus, the metric captures the robot's mechanical effort rather than the exact energy transferred to the object. For force-grounded manipulation, a natural extension is to combine joint-space work with contact detection and force/torque sensing. The success-rate drop on several tasks shows that energy regularization should be constrained by task-specific contact requirements. For example, insertion and stacking may require high work to satisfy geometric constraints. Future work should use contact-aware losses, task-conditioned weights, and energy-predictor uncertainty estimates.


\section{Conclusion}

We presented energy-regularized imitation learning for force- and work-aware manipulation. Our method defines a joint-space mechanical-work proxy from joint torque and angular displacement, learns a differentiable predictor, and uses the frozen predictor as an additional loss to fine-tune a pretrained policy. On 12 RLBench tasks, it reduces average mechanical work from 208.8 J to 204.4 J, while the mean success rate slightly increases from 86.2\% to 86.9\%. This modest reduction reflects the conservative role of the energy term, which regularizes a pretrained high-success policy without harming task performance. The results show that physically meaningful work estimates can be incorporated into imitation learning without a differentiable simulator. This direction aligns with force-grounded manipulation, where torque, contact, and mechanical work provide more deployment-relevant signals than task success alone.

\bibliographystyle{splncs04}
\bibliography{main_FGCVAM}

\end{document}